\documentclass[runningheads]{llncs}
\usepackage{graphicx}
\usepackage{subfigure}
\usepackage{amsfonts}
\usepackage{amsmath}
\usepackage{amssymb}
\usepackage{wrapfig}

\newcommand{\mcS}{\mathcal{S}}
\newcommand{\mcF}{\mathcal{F}}
\newcommand{\mcI}{\mathcal{I}}

\newcommand{\mcZ}{\mathcal{Z}}

\newcommand{\mbbR}{\mathbb{R}}
\newcommand{\mbbE}{\mathbb{E}}

\newcommand{\mbbS}{\mathbb{S}}

\newcommand{\mbfx}{\mathbf{x}}

\newcommand{\mbfxS}{\mathbf{x}^{\mathcal{S}}}

\newcommand{\bbeta}{\vec{\beta}}

\renewcommand{\and}{\mathrm{~and~}}

\newcommand{\argmax}{\arg\max}

\newcommand{\eg}{e.g.,~}
\newcommand{\ie}{i.e.,~}

\usepackage[font=small,skip=0pt]{caption}

\begin{document}
\title{Mixture Modeling of Global Shape Priors and Autoencoding Local Intensity Priors for Left Atrium Segmentation}
%
\titlerunning{Prior Modeling for Shape-Driven Left Atrium Segmentation}
%
\author{Tim Sodergren$^1$, Riddhish Bhalodia$^1$, Ross Whitaker$^1$, Joshua Cates$^2$, \\Nassir Marrouche$^2$, Shireen Elhabian$^1$}
%
\authorrunning{T. Sodergren et al.}
%
\institute{$^1$Scientific Computing and Imaging Institute, School of Computing, \\University of Utah, SLC, UT, USA  \\
\email{\{tsodergren, riddhishb, whitaker, shireen\}@sci.utah.edu}\\
$^2$ Comprehensive Arrhythmia Research and Management Center, Division of Cardiovascular Medicine, School of Medicine, University of Utah,  SLC, UT, USA\\
\email{josh.cates@gmail.com}, \email{nassir.marrouche@carma.utah.edu} }



%
\maketitle              

\begin{abstract}

Difficult image segmentation problems, \eg left atrium in MRI, can be addressed by incorporating \textit{shape} priors to find solutions that are consistent with known objects. 
Nonetheless, a single multivariate Gaussian is not an adequate model in cases with significant nonlinear shape variation or where the prior distribution is multimodal. Nonparametric density estimation is more general, but has a ravenous appetite for training samples and poses serious challenges in optimization, especially in high dimensional spaces.  
Here, we propose a maximum-a-posteriori formulation that relies on a generative image model by incorporating both local intensity and global shape priors.
We use deep autoencoders to capture the complex intensity distribution while avoiding the careful selection of hand-crafted features.
We formulate the shape prior as a mixture of Gaussians and learn the corresponding parameters in a high-dimensional shape space rather than pre-projecting onto a low-dimensional subspace. 
%
%
In segmentation, we treat the identity of the mixture component as a latent variable and marginalize it within a generalized expectation-maximization framework. 
We present a conditional maximization-based scheme that alternates between a closed-form solution for component-specific shape parameters that provides a global update-based optimization strategy, and an intensity-based energy minimization that translates the global notion of a nonlinear shape prior into a set of local penalties. 
We demonstrate our approach on the left atrial segmentation from gadolinium-enhanced MRI, which is useful in quantifying the atrial geometry in patients with atrial fibrillation. 


\end{abstract}

\vspace{-0.1in}
\section{Introduction}
\vspace{-0.1in}

Automatic image segmentation is an important enabling technology in most medical imaging applications that involve soft tissue imaging, \eg neurology, cardiology, and oncology. In particular, the preoperative anatomical representation of the left atrium (LA) is important for ablation guidance, fibrosis quantification, and biophyiscal modeling in artial fibrillation patients \cite{calkins20172017,McGann2014AtrialFA,Hansen2015AFib,Zhao2017fingerprints}. 
{\em Difficult} segmentation problems typically have two aspects.  First, \textit{image features}, such as intensity and texture are noisy and unreliable.  Second, \textit{anatomical boundaries} are ill-defined, often in the presence of low contrast, background clutter, and partial volumes, while irregular shapes with high variability limit the ability to find invariant features. Nonetheless, if one operates in a context with expectations of particular classes of anatomies (\eg LA), such challenges can be addressed by means of \textit{shape prior information} to guide and constrain the segmentation process; motivating a \textit{Bayesian formulation}. In LA segmentation context, shape-driven methods were found to be the most appropriate in addressing inherent challenges \cite{tobon2015benchmark}; thin myocardial wall, surrounding anatomical structures with similar image intensities, and topological variants pertaining to pulmonary viens arrangments \cite{ho2012left}.  In this paper, we propose a Bayesian \textit{surface-based} segmentation framework that is based on a \emph{mixture-based global shape prior} along with a \emph{feature-based local intensity prior}.

Traditionally, when dealing with low-quality image segmentation, statistical shape information has proved to be helpful in delineating correct object boundaries (\eg \cite{rousson2008prior,cremers2006kernel}). 
For example, active shape models \cite{Cootes1995} 
and their variants incorporate over-restrictive shape constraints in the form of statistical shape models (SSM) by limiting the solution to some low-dimensional linear subspace defined via training shape exemplars. However, these approaches are limited in their ability to accommodate shapes that are not represented in a low-dimensional description; a typical situation arises in applications with small and large-scale shape variability (\eg \cite{ho2012left}). Further, these methods only handle unimodal Gaussian-like shape densities.  When it comes to modeling complex shape distributions, a single Gaussian can not adequately model cases of nonlinear shape variation where the probability distribution is multimodal \cite{cremers2003shape}.

On the other end of the spectrum, multiatlas-based segmentation (MAS) approaches require a large database of atlases to capture wide range of shape variation where label maps are propagated to the testing image through registration. 
In essence, MAS can be viewed as a nonparametric estimate of the prior probabilities that converges to the true densities with the number of atlases 
\cite{awate2014multiatlas}. To reduce the computational burden introduced by registration, atlas selection is usually performed to exclude irrelevant atlases that might misguide the segmentation process \cite{aljabar2009multi}. Generally, kernel-based methods (\eg \cite{cremers2006kernel,wimmer2009generic}) 
are popular strategies for dealing with complex distributions in explicitly or implicitly represented training data. Under mild assumptions, they converge to the true distribution in the limit of infinite sample size \cite{rousson2005efficient}. For example, Cremers et al.~\cite{cremers2006kernel} used kernel density estimate (KDE) to derive non-linear shape priors which are based on a shape distance between implicitly embedded training shapes.  Nonetheless, such methods are prone to over-fitting due to small  sample size in high-dimensional shape spaces, limiting the generality of the  resulting models to fit unseen examples. Specifically, in multivariate density estimation, KDE requires larger kernel width to accommodate more exemplars as the dimension of a  variable increases. This will eventually result in model under-fit due to high bias \cite{friedman1984}. Further, for kernel based methods, the  dimension of the parameter space is proportional to the training sample size.

A finite {\em mixture} of Gaussians can approximate a full kernel density, to capture nonlinearities in the distribution or subpopulations in the underlying shape subspace \cite{cootes1999mixture} where the expectation-maximization (EM) framework 
is usually used to find the maximum likelihood estimate of mixture parameters. However, modeling shape distribution after being projected onto low-dimensional subspace
(\eg\cite{rousson2005efficient,cootes1999mixture})  
will often collapse or mix the subpopulations, which would derail learning of the mixture structure of the underlying shape space \cite{dasgupta1999learning}. Nonlinearity of shape statistics can also be modeled by lifting training shapes to a higher, probably infinite, dimensional feature (a.k.a. kernel) space where the shape distribution is assumed to be Gaussian distributed
\cite{cremers2003shape}, yet this approach results in an infinite-dimensional optimization scheme while sacrificing the efficiency of optimizing in low-dimensional subspaces \cite{rousson2005efficient}.  Further, one can settle for only an approximate solution for the reverse mapping from feature space to shape space. 

The proposed bayesian formulation relies on a maximum-a-posteriori estimation from a generative statistical model that incorporates global, nonlinear shape priors modeled as a mixture of Gaussian components, and local, nonlinear intensity prior as automatically learned image features via \textit{deep autoencoders}. 
The Gaussian mixture model (GMM) takes into account the linear subspace spanned by each mixture component to avoid the classical problem of model over fitting in high-dimensional spaces \cite{bouveyron2007high}, and can adequetely model non-linear shape variations.
To use these shape priors in segmentation, we treat the identity of the mixture component as a latent variable while marginalizing it  within a \textit{generalized expectation-maximization} framework (GEM) \cite{gelman2003bayesian}. We present a conditional maximization-based scheme that alternates between a closed-form solution for component-specific shape parameters that provides a global update-based optimization strategy, and an intensity-based energy minimization that translates the global notion of a nonlinear shape prior into a set of local penalties. Preliminary results show improved accuracy from traditional shape-based approaches.

\vspace{-0.1in}
\section{Methods}
\vspace{-0.1in}



Given an image $\mcI \in \mbbR^D$, where $D$ is the total number of voxels and $\mcI(\mbfx)$ is the intensity value of the voxel located at $\mbfx \in \mbbR^3$,  the Bayesian formulation of the segmentation problem amounts to finding the optimal surface $\mcS^*$ that maximizes the log-posterior probability $p(\mcS | \mcI)$, where a shape's surface is represented by a dense set of geometrically consistent $M-$ points (landmarks) $\{\mbfxS_i\}_{i=1}^{M}$. In order to obtain segmentations that preserve the global shape characteristics of the shape population of interest, the segmentation process is influenced by the prior, in the form of the shape probability distribution. We model such a prior distribution as a finite mixture of Gaussians,  parameterized by the mixture component identity $\mcZ$, which is treated as a  latent variable, and the model parameters $\{\Theta_z\}_{z=1}^{K}$ for $K-$components. Hence, given a latent variable $\mcZ = z$, the log-posterior can be written as\footnote{For notational simplicity, we will refer to $p(\mcZ = z) = p(\mcZ)$ hereafter.},
\begin{equation}\label{eqn:pS_I}
\log p(\mcS | \mcI) = \log p(\mcS, \mcZ = z | \mcI) - \log p(\mcZ = z | \mcS, \mcI),
\end{equation}
\noindent where $p(\mcZ | \mcS, \mcI)$ defines a conditional probability of the latent variable $\mcZ$ given the input image $\mcI$ and the estimated surface $\mcS$. 

\vspace{-0.1in}
\subsection{Generalized EM for shape-driven segmentation}
\vspace{-0.01in}

The use of log-posterior allows us to marginalize the over the latent variable $\mcZ$.  Using \textit{generalized expectation-maximization} (GEM) \cite[p.~318]{gelman2003bayesian}, we iterate to find the mode of the marginal posterior $p(\mcS|\mcI)$ by averaging over the latent variable $\mcZ$. GEM starts with an initial estimate $\mcS^o$ of the surface and iteratively refines it by marginalizing over $\mcZ$. Within each iteration, the posterior $p(\mcS | \mcI)$ is guaranteed to increase, \ie converge to a local maxima. Given the current surface guess $\mcS^{(t)}$, GEM consists of two steps: (1) \textit{E-step}, which computes the latent conditional
probability distribution $p(\mcZ | \mcS^{(t)}, \mcI) ~\forall \mcZ \in \{1,..,K\}$ and (2) \textit{M-step}, which finds a new surface $\mcS^{(t+1)}$ such that $p(\mcS^{(t+1)} | \mcI) \geq p(\mcS^{(t)} | \mcI)$. The details of the algorithm are as follows.

\vspace{0.04in}
\noindent\textbf{E-step:} The conditional probability of the latent variable $\mcZ$ given the input image and its labeling function can be computed as,
\begin{eqnarray}\label{eqn:pZ_SI}
p(\mcZ| \mcS^{(t)}, \mcI) &=& \frac{p(\mcI | \mcS^{(t)}, \mcZ) p(\mcS^{(t)}| \mcZ) p(\mcZ)}{p(\mcI | \mcS^{(t)}) p(\mcS^{(t)})}
\end{eqnarray}
\noindent where the intensity model $p(\mcI | \mcS^{(t)}, \mcZ)$ in the vicinity of the current surface guess $\mcS^{(t)}$ is assumed to be conditionally independent of the shape generating component, leaving the E-step to be fully shape-driven. Note the effect of the prior probability $p(\mcZ)$ on the expectation step, where the conditional component distribution would favor mixture components with higher support/proportion, the M-step afterwards would pull the new surface towards the most probable mixture component(s) which generated the shape to be segmented as being compatible with the local intensity and global shape priors.

\vspace{0.04in}
\noindent\textbf{M-step:} Taking the expectations on both sides of (\ref{eqn:pS_I}) while treating $\mcZ$ as a random variable with the distribution $p(\mcZ | \mcS^{(t)}, \mcI)$ yields,
\begin{eqnarray}\label{eqn:EpS_I}
\log p(\mcS | \mcI) &=& \mbbE_{p(\mcZ  | \mcS^{(t)}, \mcI)} \left[\log p(\mcS, \mcZ | \mcI)\right] - \mbbE_{p(\mcZ  | \mcS^{(t)}, \mcI)} \left[\log p(\mcZ | \mcS, \mcI)\right]
\end{eqnarray}
\noindent where the left side of (\ref{eqn:pS_I}) does not depend on $\mcZ$. The second term in (\ref{eqn:EpS_I}) is maximized when $\mcS = \mcS^{(t)}$ (the key result of EM \cite{gelman2003bayesian}). Hence, it is sufficient for the new surface estimate $\mcS^{(t+1)}$ to maximize the first term of (\ref{eqn:EpS_I}), which is the \textit{expected complete-data log-likelihood}.
\begin{eqnarray}\label{eqn:SnewMixture}
\mcS^{(t+1)} &=& \argmax_{\mcS} \sum_{\mcZ=1}^{K} p(\mcZ| \mcS^{(t)}, \mcI) 
\log p(\mcS, \mcZ | \mcI) \nonumber \\
&\propto& \argmax_{\mcS} \sum_{\mcZ=1}^{K} p(\mcZ| \mcS^{(t)}, \mcI) \left\{\log p(\mcI | \mcS, \mcZ) + \log p(\mcS|\mcZ) + \log 
p(\mcZ)\right\} 
\end{eqnarray}

\noindent The intensity term $p(\mcI | \mcS, \mcZ)$ gives rise to an intensity model in the vicinity of an estimated surface that resembles the $\mcZ-$th component. Intensity features (learned via a deep autoencoder \cite{vincent2010stacked}) $\mcF_i^{\mcI}$ are assumed to follow a normal distribution in the learned feature space with parameters $\Theta_i^{\mcF} = \{\mu_i^{\mcF}, \Sigma_i^{\mcF}\}$ that are estimated from the training data, where $\mu_i^{\mcF}$ is the mean intensity feature at the $i-$th point and $\Sigma_i^{\mcF}$ is the corresponding covariance matrix. 
\noindent  Notice that $\mcS^{(t+1)}$ increases the first term of (\ref{eqn:EpS_I}) since it is obtained based on its maximization. In addition, $\mcS^{(t+1)}$ decreases the second term of (\ref{eqn:EpS_I}), because such a term  is maximized only when $\mcS = \mcS^{(t)}$. Hence, if $\mcS^{(t+1)} = \mcS^{(t)}$, then the GEM method has converged to a local maxima of the posterior $p(\mcS | \mcI)$.  

\vspace{-0.1in}
\subsection{Mixture modeling of global shape priors}
\vspace{-0.01in}

The combination of high dimensional shape space and a small number of example shapes usually makes modeling of shape priors subject to \textit{curse-of-dimensionality}, which hinders the effectiveness of mixture learning and density estimation in high dimensional spaces. To avoid over fitting, we can parameterize the covariance matrix of a component, in a manner similar to \cite{bouveyron2007high}, through its eigen decomposition. Hence we can learn the Gaussian mixture model taking into consideration the dominant linear subspace spanned by each mixture component, where the noise variance is assumed to be isotropic and contained in a subspace orthogonal to the component's subspace.

\begin{wrapfigure}{R}{2.5in}
	\centering
	\includegraphics[width=2.5in]{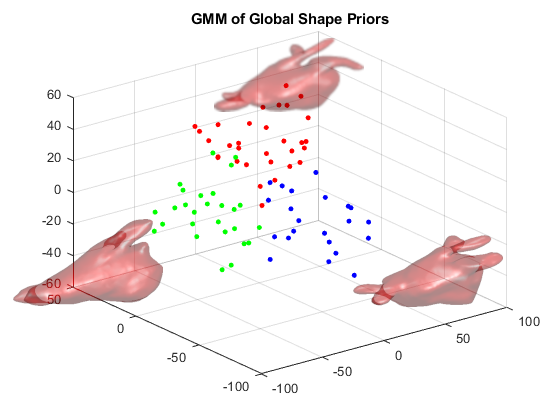}
	\caption{Low-dimensional subspace representation of high-demensional shape data. Each sample in the training set is represented by a single point with color representing different components of the GMM. An example segmentation from each component is included to illustrate differences.}
	\label{fig:gmm}
\end{wrapfigure}


We first model a dense set of $M$ homologous landmarks via particle based modeling. Assume that the anatomy-specific shape space of all shapes defined using the $M-$landmarks representation is comprised of $K-$mixture components. Each component is parameterized by $\Theta_k = \{\pi_k, \mu_k, \Lambda_k, $ $\Psi_k, d_k, \sigma_k\}$ and learned via high-dimensional EM \cite{bouveyron2007high} where $\pi_k = p(\mcZ = k)$ is the component probability/proportion, $\mu_k$ is the mean vector, $d_k$ is the intrinsic dimension of the $k-$th mixture component, i.e., the number of dominant modes of shape variation, $\Lambda_k$ is a diagonal matrix containing the largest $d_k$ eigenvalues of the component's full covariance matrix $\Sigma_k$, $\Psi_k$ is the orthonormal matrix containing the corresponding $d_k$ eigenvectors, and $\sigma_k$ is the standard deviation of the off-subspace noise. Hence the component-specific
subspace $\mbbS_k$  can be defined as follows: 
\begin{equation}
\mbbS_k = \left\{ \mcS^\beta_k = \mu_k + \Psi_k \bbeta_k~|~ 
\mcS^\beta_k \in :\mbbR^{M \times 3}, \bbeta_k \in \mbbR^{d_k} \right\}
\end{equation}

\noindent where $\bbeta_k \in \mbbR^{d_k}$ encodes the shape parameters w.r.t. the $k-$th component  subspace $\mbbS_k$.The shape probability distribution $p(\mcS)$ is defined as a weighted sum of component-conditional probabilities $\{p(\mcS | \Theta_k)\}_{k=1}^K$ using the mixture weights $\{\pi_k\}$. Notice, that the component-wise model beyond the first $d_k$ modes consists of an isotropic variance in the directions orthogonal to the subspace, which has the effect of penalizing point wise differences from the learned, $d_k$-dimensional subspace. Hence, the surface probability conditional on the $k-$th mixture component becomes a product of two marginal and indepepdent Gaussians; within-subpace and off-subspace \cite{bouveyron2007high}.

We illustrate this training step in Fig.~\ref{fig:gmm}. These data are from the samples selected for training and illustrate a low-dimensional subspace embedding of every sample colored according to their class membership in a mixture model of 3 components along with example LA segmentations from each. One can observe key differences between each component such as the number and shape of pulmonary veins.


\begin{wrapfigure}{R}{2in}
	\centering
	\includegraphics[width=2in]{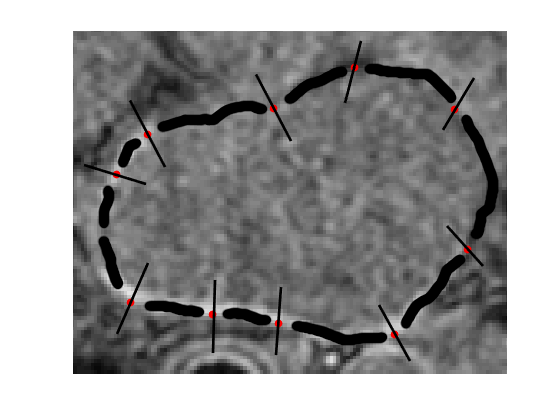}
	\caption{Intensity gradients across the segmentation surface are taken at each point location.}
	\label{fig:intensities}
\end{wrapfigure}


\vspace{-0.1in}
\subsection{Autoencoding local intensity priors}
\vspace{-0.01in}

Similar to the active shape model (ASM) approach of \cite{cootes1999mixture}, we model local intensity profiles at the object surface consisting of the normalized first derivative of image intensities oriented along the approximate normal to the surface and centered at each landmark location, with radius of $\ell$ voxels with total length of $L = 2\ell+1$ (Fig.~\ref{fig:intensities} ). Corresponding profiles (i.e., those that share the same landmarks ) are combined across all training images but are treated independently of one another. The traditional approach to segmentation is to use the Mahalanobis distance of candidate profiles to the training data as a cost function to be minimized, however, it assumes the feature space to be first order linear and normally distributed. In order to overcome this limitation, we employ deep autoencoders to adaptively learn the higher order non-linear features. In this case, we use a 2 layer $\times$ 10 hidden units sparse autoencoder to generate pseudolinear representations of the feature space for each landmark. Due to the computationally expensive nature of this step we determined these parameters through a qualitative empirical analysis but there is potential for improvement via more rigorous cross-validation. To improve the statistics and capture lateral intensity changes, we append additional profiles which are parallel to the main profile and, offset by a subvoxel distance, with interpolated intensity values, to form ``thick" profiles. We repeat this for multiple image resolutions and store the autoencoder as well as the encoded output.

\vspace{-0.1in}
\subsection{Segmentation via conditional maximization}
\vspace{-0.01in}

The marginal posterior optimization problem in (\ref{eqn:SnewMixture}) involves  two types of latent variables, the labeling function to be estimated $\mcS$ and the component-specific shape parameters $\bbeta_{k}$. To seek an optimal solution for $\mcS$, we propose 
a \textit{conditional maximization}-based scheme \cite[p. 312]{gelman2003bayesian} that alternates between two optimization phases. In the first phase, we optimize for $\{\bbeta_{k}\}_{k=1}^K$ given $\mcS$, and in the second phase we optimize for $\mcS$ given individual shape parameters $\{\bbeta_{k}\}_{k=1}^K$. Starting with the coarsest image resolution, we take the mean landmark positions from the training data as the initial model and iterate through these steps. 

\vspace{0.04in}
\noindent\textbf{Localized feature-based optimization: } For each landmark position, we generate a series of overlapping profiles of $L-$dimension that extend above and below the original profile by some specified search length $s$. For each of these candidate profiles, we generate features using pre-trained autoencoders. We then take the Mahalanobis distance of each candidate profile to the encoded training features and update the landmark position to the center of the optimal profile, oen with the smallest distance. This step ensures that landmarks are locally optimal, but they may no longer by globally optimal as there is no shape information encoded into these features. 

\vspace{0.04in}
\noindent\textbf{Component-specific shape parameters optimization:} For a given surface $\mcS$ (from the intensity-based step), component-specific shape parameters $\bbeta_k$ can be obtained by projecting the surface onto the component-specific subpace, leading a closed-form solution $\bbeta_k = \Psi_k^T(\mcS - \mu_k)$. By limiting the model parameters to what is considered a normal contour with respect to shape, the landmark positions become globally optimal. We repeat this a predetermined number of times at each resolution, coarse to fine.


\section{Results}

\begin{figure}%
	\centering
	\subfigure[MRI test image]{%
		\label{fig:mri}%
		\includegraphics[width=1.5in]{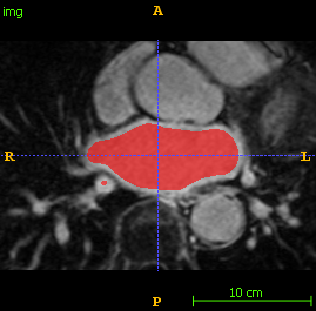}}%
	\subfigure[Original segmentation]{%
		\label{fig:rec}%
		\includegraphics[width=2in, trim=100 100 100 50, clip]{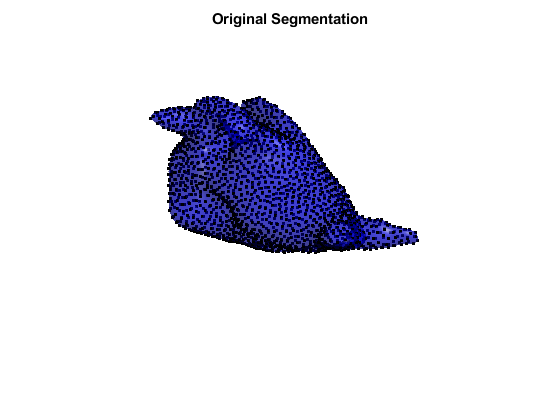}}%
	\caption{MRI image (a) of test sample next to LA segmentation with landmarks (b).}
	\label{fig:testimage}
\end{figure}

\begin{figure}
	\centering
	\includegraphics[width=\linewidth]{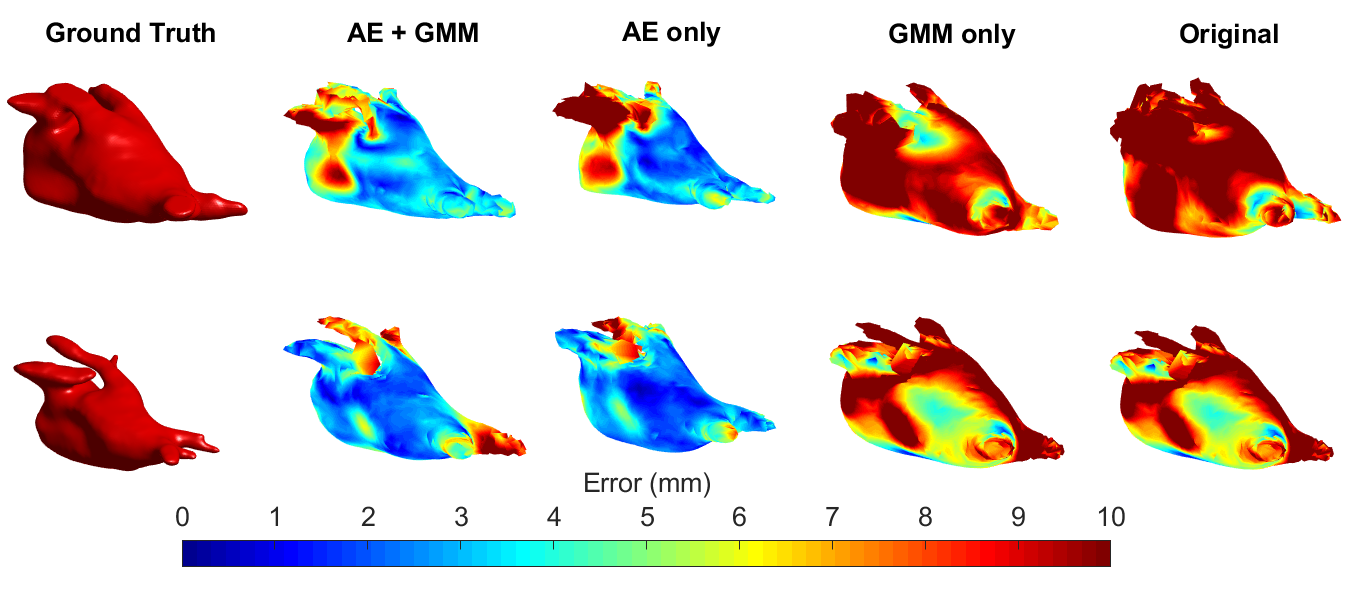}
	\caption{Example results for 2 different samples with point by point error mapped onto segmentation surface. AE = autoencoder, GMM = mixture modeling, Original = neither (original ASM method).}
	\label{fig:results}
\end{figure}

We tested these methods on 100 3D MRI images, separated into 80 for training and 20 for testing. Fig.~\ref{fig:testimage} shows a sample MRI image with the provided segmentation. The segmented volume is also separately displayed in blue along with modeled landmark positions. We first develop a shape model of 2048 homologous landmark positions modeled with a 3 component Gaussian mixture model. We then model local intensity gradients for each position across all samples using an autoencoder on profiles 11 voxels in length (optimized through cross-validation). In order to improve model statistics, we encode additional parallel information to create "thick" profiles. This model was then used to segment the 20 test images according to the methods described above. To compare the effectiveness of our method we evaluate results both with and without autoencoded local intensity profiles as well as with and without the mixture model. Results are in Table \ref{table:dice} .
While there appears to be some improvement in median Dice coefficient, this may not be the best metric for evaluating the efficiency of the proposed method \cite{Taha2015}. We also evaluated the euclidean distance between final landmark positions from our segmentations to those of the ground truth. 
This also allows for a more detailed point by point analysis of error as shown in Fig.~\ref{fig:results}. Here we see two examples from the test data set showing clear improvement in segmentation results when using autoencoded profiles. We truncate the color scale at 10 (1 cm) and flag any higher error as an extreme outlier. Error is generally limited to image edges around the pulmonary veins when using autoencoded intensity profiles as opposed to nearly universally high error without. 
These results could be further improved by initializing the model with the component mean shapes as opposed to the global mean which is how we are presently doing it.

\begin{table}[h!]
	\caption{Dice scores for test results.}
	\label{table:dice}
	\centering
	\begin{tabular}{|c|c|c|c|c|}
		\hline 
		\rule[-1ex]{0pt}{2.5ex} Autoencoder & yes & yes & no & no \\ 
		\hline 
		\rule[-1ex]{0pt}{2.5ex} Mixture Model & yes & no & yes & no \\ 
		\hline 
		\rule[-1ex]{0pt}{2.5ex} Median Dice Coefficient & 0.743 & 0.747 & 0.729 & 0.739 \\ 
		\hline 
	\end{tabular}
\end{table}


\section{Conclusion}
This paper proposed a shape/feature-based generative model for left atrium segmentation by modeling non-Gaussian global shape priors as mixture of Gaussians on landmark-based representations of training data and learning feature-based representations for local intensity priors. The mixture parameters were learned in the high dimensional shape space by taking into account component-specific subspaces to avoid over fitting. The method used a variant of ASM-based segmentation framework that relies on a maximum a-posteriori estimation with a marginalization over class membership within a generalized EM framework. While these results are preliminary, they suggest that ASM’s for LA segmentation can be improved through optimized prior modeling. Local optimization through autoencoding of non-linear intensity features and global optimization through mixture modeling of shape parameters increases the accuracy of the original method

\vspace{0.1in}
\noindent\textbf{Acknowledgment:} This work was supported by the National Institutes of Health [grant numbers R01-HL135568-01 and P41-GM103545-19]. 

%
\vspace{-0.1in}
{\small 
	\bibliographystyle{splncs04}
	\bibliography{references}
}

\end{document}